\documentclass[10pt,journal,compsoc]{IEEEtran}
\usepackage{ifpdf}

\ifCLASSOPTIONcompsoc
  \usepackage[nocompress]{cite}
\else
  \usepackage{cite}
\fi
\usepackage{amssymb}
\usepackage{caption, subcaption}
\usepackage{xspace}
\usepackage{multirow}
\usepackage{color}
\usepackage{enumitem}

\newcommand{\ie}{\emph{i.e.},\xspace}
\newcommand{\eg}{\emph{e.g.},\xspace}

\def\B#1{\mathbf #1}
\def\C#1{\mathcal #1}
\ifCLASSINFOpdf
  \usepackage[pdftex]{graphicx}
\else
  \usepackage[dvips]{graphicx}
\fi
\usepackage{amsmath}

\begin{document}
%
\title{Hierarchical Aspect-guided Explanation Generation for Explainable Recommendation}
%
%
%
%

\author{Yidan Hu, Yong Liu, Chunyan Miao, Gongqi Lin, Yuan Miao
\IEEEcompsocitemizethanks{\IEEEcompsocthanksitem Yidan Hu and Chunyan Miao are currently with School of Computer Science and Engineering, Nanyang Technological University, Singapore 639798. Email: yidan001@e.ntu.edu.sg, ascymiao@ntu.edu.sg.
\IEEEcompsocthanksitem Yong Liu is with Joint NTU-UBC Research Centre of Excellence in Active Living for the Elderly (LILY), and AlibabaNTU Singapore Joint Research Institute, Nanyang Technological University, Singapore 639798. Email: stephenliu@ntu.edu.sg.

\IEEEcompsocthanksitem Yuan Miao and Gongqi Lin are currently with IT Discipline, Victoria University, Australia. Email: yuan.miao@vu.edu.au, lingongqi2009@gmail.com}
\thanks{Manuscript received XXX; revised XXX.}}

%
%

\markboth{IEEE Transactions on Knowledge and Data Engineering}%
{Shell \MakeLowercase{\textit{et al.}}: Bare Demo of IEEEtran.cls for Computer Society Journals}
%



\IEEEtitleabstractindextext{%
\begin{abstract}
Explainable recommendation systems provide explanations for recommendation results to improve their transparency and persuasiveness. 
The existing explainable recommendation methods generate textual explanations without explicitly considering the user's preferences on different aspects of the item. In this paper, we propose a novel explanation generation framework, named \textbf{\underline{H}}ierarchical \textbf{\underline{A}}spect-guided explanation \textbf{\underline{G}}eneration (\textbf{HAG}), for explainable recommendation. 
Specifically, HAG employs a review-based syntax graph to provide a unified view of the user/item details. 
An aspect-guided graph pooling operator is proposed to extract the aspect-relevant information from the review-based syntax graphs to model the user's preferences on an item at the aspect level. Then, a hierarchical explanation decoder is developed to generate aspects and aspect-relevant explanations based on the attention mechanism. 
The experimental results on three real datasets indicate that HAG outperforms state-of-the-art explanation generation methods in both single-aspect and multi-aspect explanation generation tasks, and also achieves comparable or even better preference prediction accuracy than strong baseline methods.
\end{abstract}

}

\maketitle

\IEEEdisplaynontitleabstractindextext

%
\IEEEpeerreviewmaketitle

\IEEEraisesectionheading{\section{Introduction}\label{sec:introduction}}

%
%
%
%
\IEEEPARstart{R}{ECOMMENDATION} systems have been widely used to help users make decisions by suggesting them a list of items they may have interests. Various types of recommendation methods have been developed based on  collaborative filtering~\cite{shi2014collaborative} and deep learning techniques~\cite{zhang2019deep}. Although these methods can usually achieve satisfactory performances, it is still very hard to explain their recommendation mechanisms. Thus, many recent research efforts~\cite{catherine2017transnets,chen2018neural,chen2019generate,chen2019personalized,sun2020dual} have been devoted to building explainable recommendation models that explain why an item is recommended by generating high-quality explanations, which can help improve the transparency and persuasiveness of recommendation systems. 
In practice, different strategies may be adopted to explain the recommendation results, \eg images, the behaviors of relevant users, and textual descriptions of relevant items~\cite{zhang2018explainable}. In this work, we focus on generating high quality review text to explain the recommendation results presented to the user.

The review generation methods for explainable recommendation can be roughly classified into two groups:
template-based approaches and natural language generation approaches~\cite{zhang2018explainable}. The template-based methods generate explanations by filling the generated words in a sentence template. For example, in the template ``\textit{You might be interested in [\textbf{\textit{aspect}}], on which this product performs well}'', we can replace \textit{[\textbf{\textit{aspect}}]} by a generated aspect to produce text explanation for item recommendation~\cite{zhang2014explicit}. However, template-based explanations are uninformative and not persuasive. Moreover, designing high-quality templates usually requires domain knowledge. Natural language generation approaches can generate more natural and flexible sentences. Such approaches have recently attracted increasing research attention. 
However, the pioneer works~\cite{dong2017learning,li2017neural} focus on generating short reviews or tips only based on given attributes (\textit{e.g.,} user ID, item ID, and rating value). Thus, it is difficult for them to generate reliable explanations without considering other generative signals~\cite{zhang2014explicit,li2017neural}.

Aspects, an important type of generative signal, which usually represent item features (\textit{e.g.,} "price"  and "romance"), have recently been exploited to build aspect-aware explanation generation models~\cite{ni2018personalized,ni2019justifying,li2020generate}. 
In these methods, the aspects are extracted from the user-generated reviews and used to train the explanation generation model. 
These methods assume the user's interested aspects are available for explanation generation. In practice, this assumption usually does not hold, because we need to predict the user's preferences for different aspects of a target item in many application scenarios. The hierarchical generation framework~\cite{li2019generating} provides a potential solution to address this problem. However, this method only considers the  user ID, item ID, and the rating value as inputs. Thus, it 
cannot be effective in capturing the aspect-relevant details of the user and item for generating long and informative explanations.

\begin{figure*}
\centering
\includegraphics[width=0.95\textwidth]{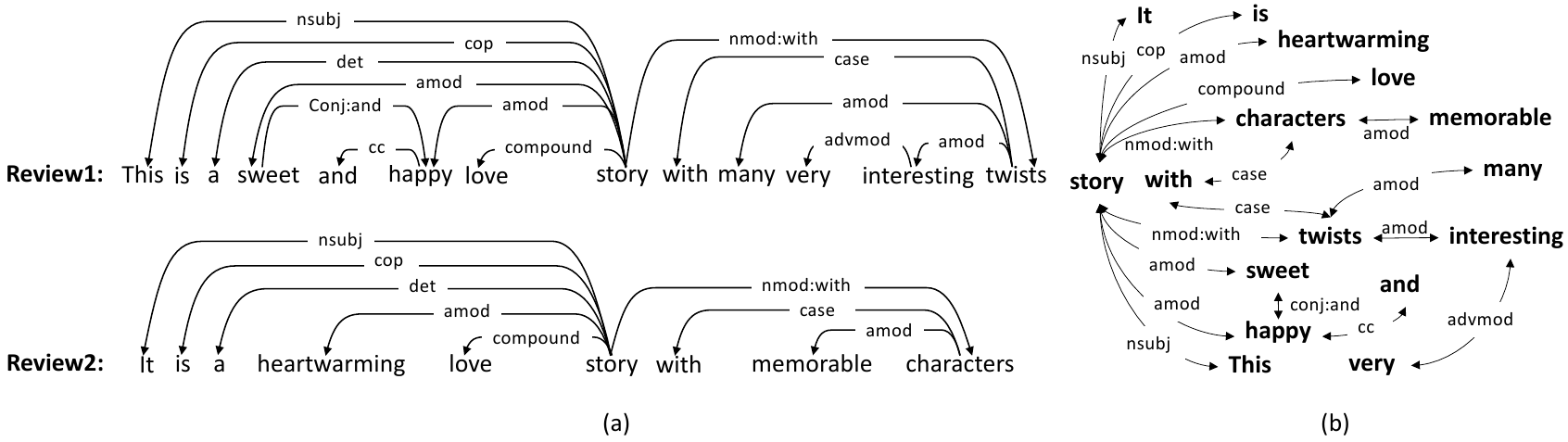} 
\caption{An example of the review-based syntax graph. (a) shows the dependency tree structures of two reviews. (b) shows the review-based syntax graph built based on the nodes and relations extracted from the dependency trees of reviews.}
\label{figure:semanticgraph}
\end{figure*}

To address this problem, we build a review-based syntax graph to provide a unified view of the user/item details based on the review data. Firstly, a syntax dependency tree is built from each review.
The relations in the dependency tree provide important clues to mine aspects, details, and opinions. From Figure~\ref{figure:semanticgraph} (a), we observe two sub-trees: \textit{story} $\xrightarrow[]{\text{nmod:with}}$ \textit{twists} $\xrightarrow[]{\text{amod}}$ \textit{interesting}, \textit{story} $\xrightarrow[]{\text{nmod:with}}$ \textit{characters} $\xrightarrow[]{\text{amod}}$ \textit{memorable}. They are both built up with the structure of \textit{aspect $\xrightarrow[]{}$ details $\xrightarrow[]{}$ opinion}. 
To aggregate the details of the same aspect in different reviews, we construct the review-based syntax graph by connecting details in different dependency trees. As shown in Figure~\ref{figure:semanticgraph} (b), details about \textit{love story}, such as \textit{characters} and \textit{twists}, can be directly connected to \textit{love story} in the review-based syntax graph.

Moreover, we propose a novel explainable recommendation framework, \textit{i.e.,} \underline{H}ierarchical \underline{A}spect-guided explanation \underline{G}eneration (HAG). Specifically, HAG performs hierarchical aspect-guided graph pooling on the user/item review-based syntax graph to extract the aspect-relevant information for building the user/item representation. The user's interests are matched with the item properties at the aspect level to predict her preferences for different aspects of the item. Then, a hierarchical explanation decoder is developed to generate aspects and aspect-relevant explanations based on attention mechanisms. To demonstrate the effectiveness of the proposed HAG model, we perform extensive experiments on three real-world datasets. The experimental results indicate that HAG outperforms state-of-the-art explanation generation methods, and achieves comparable or even better preference prediction accuracy than baseline methods.

\begin{figure*}
\centering
\includegraphics[width=0.95\textwidth]{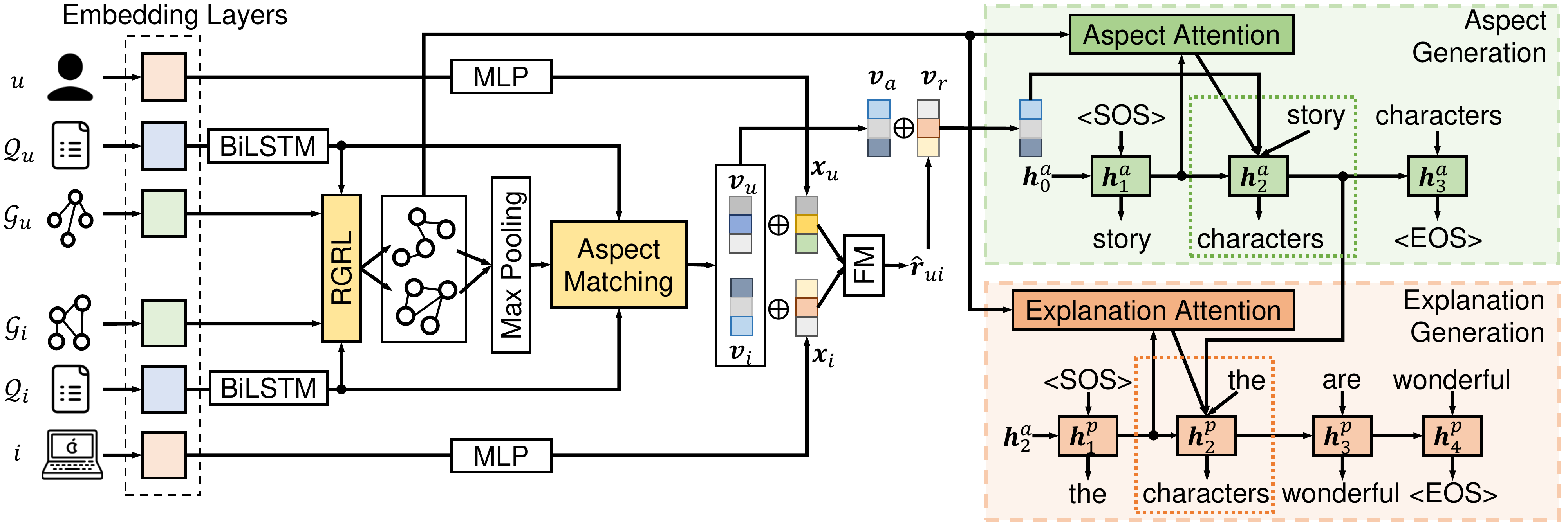}
\caption{Overall structure of the proposed HAG model. In this figure, we use the aspect and explanation generation processes of a single word ''characters" as an example.}
\label{fig:framework}
\end{figure*}

\section{Related Work}
There are two main groups of explanation generation methods for explainable recommendation, namely template-based and generation-based methods.

Template-based methods generate recommendation explanations by filling pre-defined templates with different words for different users. For example, the explicit factor model~\cite{zhang2014explicit} generates the explanations by filling in the aspect in the pre-defined template. Moreover, \cite{wang2018explainable} introduces a template-based explainable recommendation model with aspects and opinions. However, these methods cannot provide more details about user preferences, and manually designing the template is also time-consuming.

Generation-based methods focus on developing natural language generation methods for explainable recommendation. 
For example, \cite{dong2017learning,li2017neural,tang2016context} employ Recurrent Neural Networks (RNNs) based methods to encode the attribute information, \textit{e.g.,}  user ID, item ID, and rating value, to generate explanations for recommendation results. These methods cannot generate reliable and precise explanations due to the lack of guided information. To address this problem,
\cite{ni2018personalized} exploits auxiliary information to generate explanations from given aspects. Moreover, \cite{li2020generate} proposes the neural template-based explanation generation framework, which integrates the advantages of both the template-based and language generation methods. This method first uses the given aspect as a template and then generates template-controlled explanations.

The aforementioned methods usually focus on short explanation generation with only one aspect. 
To support long and informative explanation generation, \cite{li2019generating} proposes a coarse-to-fine generation framework that first generates a sentence skeleton and then generates the aspect-aware explanation. This method only considers the user ID, item ID, and rating value as input attributes. As these input attributes usually contain less information, \cite{ni2019justifying} introduces the reference-based seq2seq model that treats historical justifications as references. 
In addition, \cite{li2020knowledge} presents a knowledge enhanced review generation model to combine review and knowledge graph information.

\section{PRELIMINARY}
In this section, we introduce some background about the construction of review-based syntax graph and the research problem studied in this work.

\subsection{Review-based Syntax Graph Construction}
\label{sec:graph_build}
In this work, we denote the historical reviews of a user $u$ by $\C{D}_u = \{d_u^1,d_u^2,\cdots, d_u^{n_{du}}\}$ and the historical reviews of an item $i$ by $\C{D}_i = \{d_i^1,d_i^2,\cdots,d_i^{n_{di}}\}$, where $n_{du}$ and $n_{di}$ denote the number of reviews associated with $u$ and $i$, respectively. 
To better understand the review data, we build a user review-based syntax graph $\C{G}_u$ and an item review-based syntax graph $\C{G}_i$ from the review sets $\C{D}_u$ and $\C{D}_i$, respectively. For each user $u$, we first apply text pre-processing techniques (\eg tokenization and spelling correct) on $\C{D}_u$. Then, dependency parsing is used to automatically generate constituent-based representation (\ie dependency tree) for each review sentence based on syntax~\cite{jurafsky2018speech}. Moreover, we also perform pruning to remove the words with little information. For example, as "det"  in Figure 1 (a) in the main content is the relation between "story" and its determiner "a" which has little semantic information, thus we remove the relation "det" and only keep the head and tail nodes. Similarly, we also remove the relations such as "nmod:poss" and "punct". Then, we connect all the dependency trees of the review sentences in $\C{D}_u$ by replacing the same words in different dependency trees with the same node. After that, we can obtain the user review-based syntax graph $\C{G}_u=\{\C{X}_u, \C{E}_u\}$, where $\C{X}_u$ denotes the set of nodes (\ie words), and $\C{E}_u$ denotes the set of edges $\C{E}_u = \{(x_h, r, x_t)|x_h, x_t \in \C{X}_u, r \in \C{R}\}$. Here, $r$ denotes the relation connecting the two nodes, and $\C{R}$ denotes the set of all possible relations. Similar operations are performed on the item reviews $\C{D}_i$ to obtain the item review-based syntax graph $\C{G}_i=\{\C{X}_i, \C{E}_i\}$.
\subsection{Problem Formulation}

Following~\cite{he2015trirank,li2020generate}, we first extract aspects from the observed review data using the tool developed in~\cite{zhang2014explicit}, and use $\C{L}_u$ and $\C{L}_i$ to denote the lists of aspects derived from the user review set $\C{D}_u$ and item review set $\C{D}_i$, respectively. 
In $\C{L}_u$, we sort aspects according to their occurrence frequency in descending order, and choose the top-$n$ ranked aspects to describe the user properties, which are denoted by $\C{Q}_u = \{q_u^1,q_u^2,\cdots,q_u^{n}\}$. Similarly, we choose the top-$n$ ranked aspects from $\C{L}_i$ and denote them by $\C{Q}_i = \{q_i^1,q_i^2,\cdots,q_i^{n}\}$.
In this work, we study the following explainable recommendation problem: \textit{given a user $u$ and an item $i$, their historical aspect sets $\C{Q}_u$ and $\C{Q}_i$, and review-based syntax graphs $\C{G}_u$ and $\C{G}_i$, we aim to predict the user's preference $r_{ui}$ on the item, mine the user's interested aspects $\C{A_{ui}} = \{a_1,a_2,\cdots,a_{m}\}$ of the item, and generate aspect-aware explanations $\C{P_{ui}} =\{ p_1,p_2, \cdots, p_m\} $, in the form of a set of sentences}.

\section{The Proposed HAG Framework}
Figure~\ref{fig:framework} shows the overall framework of the proposed explainable recommendation framework HAG. As shown in Figure~\ref{fig:framework}, HAG contains the following main components: 1) review-based syntax graph representation learning (RGRL), 2) aspect matching, 3) preference prediction, and 4) hierarchical explanation decoder. Next, we introduce the details of each component.

\subsection{Review-based Syntax Graph Representation Learning}
The objective of the RGRL module is to extract representations for each review-based syntax graph. Firstly, we apply a BiLSTM $f$ to encode the word embedding of an aspect $q$, and obtain output backward hidden state as the representation $f(q)$. Then, hierarchical aspect-guided graph pooling is used to extract the review-based syntax graph representation at the aspect level. We propose an aspect-guided graph pooling (AGP) operator to effectively extract the aspect-specific knowledge from the review-based syntax graph. AGP employs the user/item historical aspects to guide review-based syntax graph representation learning.

\subsubsection{Aspect-guided Graph Pooling Operator}
\label{sss:agp}

\begin{figure}
\centering
\includegraphics[width=0.95\columnwidth]{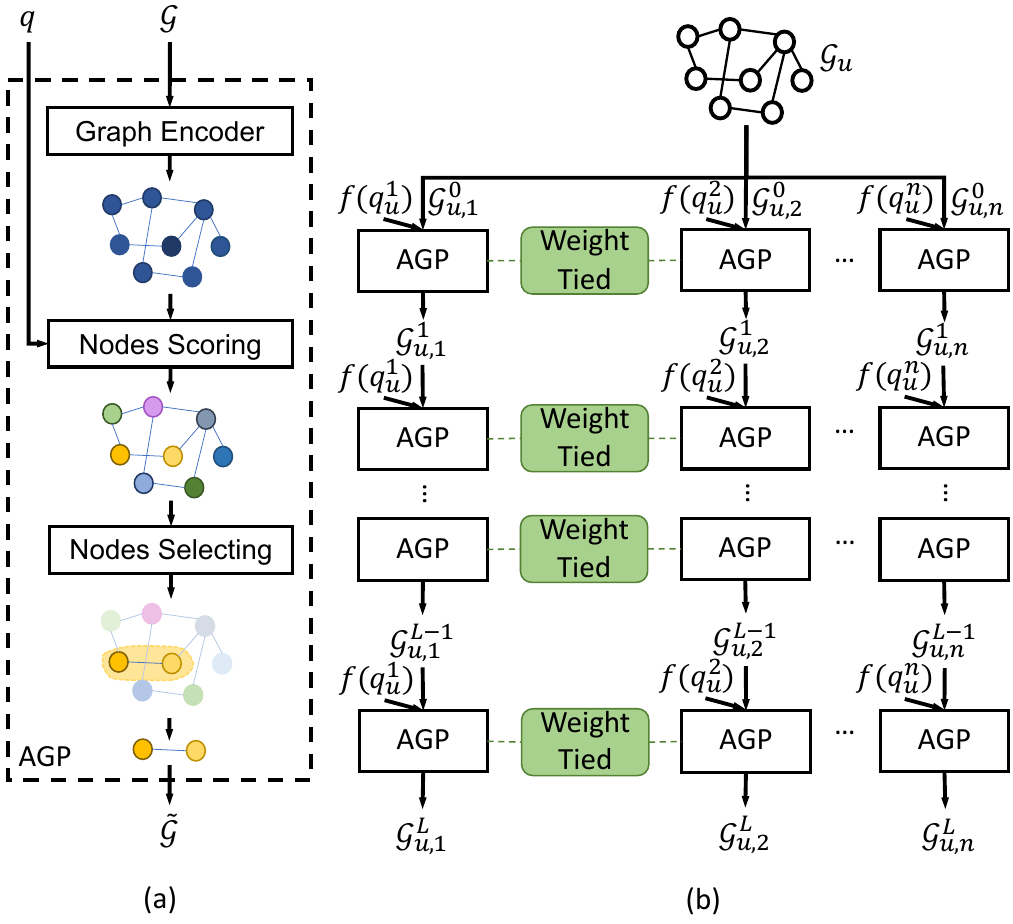}
\caption{(a) shows the flow of the AGP operator where $q$ and $\C{G}$ denote the input aspect and graph. (b) shows the flow of review-based syntax graph representation learning.}
\label{fig:AGAN}
\end{figure}

Figure~\ref{fig:AGAN} (a) shows the workflow of the AGP operator. The inputs of an AGP operator include an input graph $\C{G}=\{\C{X}, \C{E}, \B{X}, \B{A}\}$ and an aspect $q$ with its representation ${f(q)}$. Here, $\C{X}$ and $\C{E}$ denote the set of nodes and edges in $\C{G}$, respectively. $\B{X}$ is the node feature matrix and $\B{A}$ is the adjacency matrix. Firstly, a graph attention network (GAT)~\cite{velivckovic2017graph} is used to encode graph $\C{G}$. For each node $x_h \in \C{X}$, we denote its first-hop neighbors in the graph by $\C{N}_{h}$. The feature of $x_{h}$ is updated by aggregating the input features of neighborhood nodes and adding its input feature $\B{x}_{h}$ by self-loop as,
\begin{equation}\label{equ:dp}
\overline{\B{x}}_{h} = \B{x}_{h}\ + \sum_{{x}_t \in \C{N}_{h}} \alpha(x_h, r,x_t)\B{x}_{t}\B{W}_1,
\end{equation}
where $\B{W}_1 \in \mathbb{R}^{d_0 \times d_0}$ is the weight matrix, and $\alpha(x_h,r,x_t)$ denotes the attention score between two nodes $x_t$ and $x_h$.
Following~\cite{velivckovic2017graph}, we define the attention score $\alpha(x_h,r,x_t)$ as,
\begin{align}\label{equ:dp1}
\alpha(x_h,r,x_t)=\frac{\exp\big[\pi(x_h,r,x_t)\big]}{{\sum_{{x}_{t'}\in \C{N}_{h}}}\exp\big[\pi(x_h,{r},{x}_{t'})\big]}.
\end{align}
Here, $\pi(x_h,r,x_t)$ is implemented by the following attentional mechanism,
\begin{equation}\label{equ:dp2}
\pi(x_h,r,x_t)=\sigma_1\big((\B{x}_{h}\B{W}_{2})(\B{x}_{t}\B{W}_{3}+\B{r}\B{W}_{4})^{\top}\big),
\end{equation}
where $\B{r}$ is the embedding of the relation $r$, $\sigma_1(\cdot)$ is LeakyReLU activation function, $\B{W}_{2}$, $\B{W}_{3}$, $\B{W}_{4}\in \mathbb{R}^{d_0 \times d_0}$ are weight matrixes. We use a matrix $\overline{\B{X}}$ to denote the updated features of all nodes.

Inspired by previous graph pooling methods~\cite{gao2019graph,gao2019learning,lee2019self}, we define the following aspect-aware importance score to describe the relevance of each node in $\C{G}$ to the given aspect $q$,
\begin{equation}\label{equ:mse}
\delta(x_h) =\mbox{abs}\big(\overline{\B{x}}_{h}{f(q)}^{\top}\big),
\end{equation}
where $\mbox{abs}(\cdot)$ denotes the absolute value function.
The nodes in $\C{G}$ can be ranked according to $\delta(x_h)$ in descending order. Then, we denote the set of top-$K$ ranked nodes by $\widetilde{\C{X}}$ and their indices by $\mbox{idx}(s)$. In this work, we empirically set $K=\lceil \rho |\C{X}|\rceil$, where $\rho$ is the pooling ratio, $|\cdot|$ denotes the cardinality of a set, $\lceil \cdot \rceil$ is the ceiling function. The new features of nodes in $\widetilde{\C{X}}$  and adjacency matrix $\widetilde{\B{A}}$ of the corresponding graph are defined as,
\begin{gather}
\widetilde{\B{X}} =  \mbox{ReLU}\big(\overline{\B{X}}[\mbox{idx}(s),:] \B{W}_5+\B{b}_1\big),\\
\widetilde{\B{A}} = \B{A}\big[\mbox{idx}(s), \mbox{idx}(s)\big],
\end{gather}
where $\B{W}_5\in \mathbb{R}^{K \times d_0}$ and $\B{b}_1\in \mathbb{R}^{1 \times d_0}$ are the weight matrix and bias vector respectively. $\overline{\B{X}}[\mbox{idx}(s),:]$ is row-wise indexed feature matrix. $\B{A}[\mbox{idx}(s),\mbox{idx}(s)]$ aims to get the row-wise and column-wise indexed adjacency matrix from $\B{A}$. Then, $\widetilde{\B{X}}$ and $\widetilde{\B{A}}$ are the new feature matrix and the corresponding adjacency matrix after pooling. Moreover, we use $\widetilde{\C{E}}$ to denote the set of edges that describe the connecting relationships between the nodes in $\widetilde{\C{X}}$. Then, the output of the AGP operator is denoted by $\widetilde{\C{G}}=\{\widetilde{\C{X}}, \widetilde{\C{E}}, \widetilde{\B{X}}, \widetilde{\B{A}}\}$.

\subsubsection{Hierarchical Graph Pooling}
We perform hierarchical graph pooling on the item/user review-based syntax graph, by staking AGP operators. 
Here, we only introduce the hierarchical graph pooling on the user review-based syntax graph $\C{G}_u$. The same process is also followed on the item review-based syntax graph $\C{G}_i$. As shown in Figure~\ref{fig:AGAN} (b), we conduct $L$ layers graph pooling on $\C{G}_u$, guided by the user aspect set $\C{Q}_u = \{q_u^1,q_u^2,\cdots,q_u^{n}\}$.

At the $\ell$-th layer, there are $n$ AGP operators. The input graph of the $k$-th AGP operator is $\C{G}_{u,k}^{\ell-1}=\{\C{X}_{u,k}^{\ell-1}, \C{E}_{u,k}^{\ell-1}, \B{X}_{u,k}^{\ell-1}, \B{A}_{u,k}^{\ell-1}\}$. The representation ${f({q}}_u^k)$ of aspect $q_u^k$ is used to guide the pooling in the $k$-th AGP operator. The output graph of this AGP operator is $\C{G}_{u,k}^{\ell}$, which will be used as the input of the $k$-th AGP operator at the $(\ell+1)$-th layer. Note that, at the first layer, the inputs $\C{G}_{u,k}^{0}=\C{G}_u$, for $k=1, 2, \cdots, n$.
We perform max pooling on $\B{X}_{u,k}^\ell$ to obtain the aspect-aware graph representation $\B{g}_{u,k}^{\ell}$ at the $\ell$-th layer. After performing the above pooling operation $L$ times, we can obtain multiple representations of $\C{G}_u$ that are relevant to the aspect $q_u^k$, \textit{i.e.,} $\B{g}_{u,k}^{1}, \B{g}_{u,k}^{2}, \cdots, \B{g}_{u,k}^{L}$. To fuse the graph representations from fine-grained to coarse, we concatenate these representations to form the representation $\B{g}_u^k$ of the review-based syntax graph $\C{G}_u$ as $\B{g}_u^k=\B{g}_{u,k}^1\oplus\B{g}_{u,k}^2\oplus\cdots \B{g}_{u,k}^L$.

\subsection{Aspect Matching}
To better describe the user's preferences for different aspects, we compute aspect level user representation ${\B{S}}_u$ and item representation ${\B{S}}_i$ by concatenating the representations of historical aspects with those of the review-based syntax graphs as follows, 
\begin{equation}
{\B{S}}_u=
\begin{bmatrix}
\B{g}_u^1 \oplus f(q_u^1) \\
\B{g}_u^2 \oplus f(q_u^2) \\
\cdots\\
\B{g}_u^{n} \oplus f(q_u^{n})
\end{bmatrix},{\B{S}}_i=\begin{bmatrix}
\B{g}_i^1 \oplus f(q_i^1) \\
\B{g}_i^2 \oplus f(q_i^2) \\
\cdots\\
\B{g}_i^{n} \oplus f(q_i^{n})
\end{bmatrix}.
\end{equation}
Following~\cite{chin2018anr,lu2016hierarchical}, we use ${\B{S}}_u$ and ${\B{S}}_i$ as features to define the following aspect level importance weight matrix,
\begin{equation}\label{equ:mse}
    \B{M}_s = \mbox{ReLU}\big({\B{S}}_u \B{W}_s {\B{S}}_i^{\top}\big),
\end{equation}
where $\B{W}_s\in \mathbb{R}^{d'\times d'}$ is a learnable weight matrix, and $d' = d_0\times L + d_1$. In $\B{M}_s\in \mathbb{R}^{n\times n}$, each element $\B{M}_s[x, y]$ describes the importance of the $y$-th item aspect to the $x$-th user aspect. Then, we fuse the aspect level information of the user and item with $\B{M}_s$ as follows,
\begin{align}
 \B{v}_u &= \phi \bigg({\B{S}}_u\B{W}_u^s \oplus \B{M}_s({\B{S}}_i\B{W}_i^s)\bigg),\\
 \B{v}_i &= \phi \bigg({\B{S}}_i\B{W}_i^s \oplus \B{M^{\top}}_s({\B{S}}_u\B{W}_u^s)\bigg),
\end{align}
where $\B{W}_u^s$, $\B{W}_i^s \in \mathbb{R}^{n \times d'}$ are trainable parameters, and $\phi(\cdot)$ denotes the mean pooling operation. 
Here, $\B{v}_u$ aims to incorporate user interested aspects and the user preferences for the item aspect. Similarly, $\B{v}_i$ aims to combine item typical aspect and the representation of aspects that are highly relevant to the user.

\subsection{Preference Prediction}
For each user $u$, we feed the  user ID embedding $\B{e}_u$ into a Multi-Layer Perceptron (MLP) and concatenate its output with her aspect level preference vector $\B{v}_u$ to form the final representation as follows,
\begin{equation}
\B{x}_u =\mbox{MLP}_u(\B{e}_u)\oplus \B{W}_u^v\B{v}_u,
\end{equation}
where $\B{W}_u^v\in \mathbb{R}^{(2d')\times d_2}$ is a learnable parameter.
Similarly, we can obtain the final representation $\B{x}_i$ of the item $i$. Then, we concatenate $\B{x}_u$ and $\B{x}_i$ to form $\B{x}$.
A Factorization Machine (FM) layer \cite{rendle2010factorization} is applied to predict the user $u$'s preference on the item $i$ as follows,
\begin{equation}\label{equ:rating_p}
\hat{r}_{ui} = b_0 +b_u +b_i +\B{x}\B{w}^\top +\sum_{ i=1}^n\sum_{j=i+1}^n\left \langle \B{v}_i,\B{v}_j \right \rangle x_ix_j,
\end{equation}
where $b_0$, $b_u$ and $b_i$ are global bias, user bias, and item bias. $\B{v}_i$ and $\B{v}_j$ are the $i$-th and $j$-th variants. $\B{w}\in \mathbb{R}^{1\times (4d_2)}$ is coefficient vector, and $\left \langle\cdot,\cdot \right \rangle $ is the dot product of two vectors. 

\subsection{Hierarchical Explanation Generation}
In HAG, we adopt two attention-based Long Short-Term Memory (LSTM) models as the aspect decoder and explanation decoder, respectively. This section introduces the details of these two decoders. 

\subsubsection{Aspect Decoder}
\label{sssec:aspect_decoder}
As shown in Figure \ref{fig:AGAN}, for each aspect $q_u^k$, we can obtain the node feature matrix $\B{X}_{u,k}^L$ of the graph $\C{G}_{u,k}^{L}$ after graph pooling on the user review-based syntax graph. 
Similarly, we can also obtain the node feature matrix $\B{X}_{i,k}^L$ after graph pooling on the item review-based syntax graph. Then, we stack all aspect-relevant node feature matrices to form the following matrices,
\begin{equation}
\B{X}_{u}=
\begin{bmatrix}
\B{X}_{u,1}^L \\
\cdots\\ 
\B{X}_{u,n}^L
\end{bmatrix},\quad 
\B{X}_{i}= 
\begin{bmatrix}
\B{X}_{i,1}^L \\
\cdots\\
\B{X}_{i,n}^L
\end{bmatrix}.
\end{equation}
To initial the hidden state, we first map the predicted rating $\hat{r}_{ui}$ into a sentiment representation $\B{v}_r$ to guide aspect and explanation generation as, 
\begin{align}
\B{v}_r = \mbox{ReLU}(\B{W}_u^r \hat{r}_{ui}+\B{b}_u^r),
\end{align}
where $\B{W}_u^r\in \mathbb{R}^{1\times d_1}$ and $\B{b}_u^r\in \mathbb{R}^{1\times d_1}$ are a trainable weight matrix and a bias vector. Then, we feed $\B{v}_r$, $\B{v}_u$, and $\B{v}_i$ into an MLP layer as,
\begin{align}
\B{h}_0^a &= \mbox{MLP}_h(\B{v}_r\oplus \B{v}_u\oplus \B{v}_i).
\end{align}
To fully exploit the review-based syntax graph information, at each decoding time-step $j-1$,  we incorporate hidden state $\B{h}_{j-1}^a$ and node feature $\B{x}_z\in \B{X}_{u}$ to calculate attention vector $\B{c}_{u,j-1}^{a}$ as,
\begin{align}
     \beta_{u}^{j-1}& = \frac{\exp(\mbox{MLP}(\B{x}_z\oplus \B{h}_{j-1}^a))}{\sum_{\acute{\B x}_{z} \in \B{X}_{u}}\exp(\mbox{MLP}(\acute{\B x}_{z}\oplus \B{h}_{j-1}^a)},\nonumber\\
     \B{c}_{u,j-1}^{a}& = \sum_{\B{x}_z \in \B{X}_{u}}  \beta_{u}^{j-1}\B{x}_z.\label{equ:att_aspect}
\end{align}
Similarly, we can obtain the attention vector $\B{c}_{i,j-1}^{a}$ from $\B{X}_i$. The next time-step hidden state $\B{h}_{j}^a$ is,
\begin{align}\label{equ:mse}
\hat{\B{h}}_{j-1}^a &= \B{h}_{j-1}^a\oplus \B{c}_{u,j-1}^{a}\oplus \B{c}_{i,j-1}^{a} \oplus  \B{h}_{0}^a, \nonumber\\
\B{h}_{j}^a &= \mbox{LSTM}(\hat{\B{h}}_{j-1}^a,E(w_{j-1}^a)),
\end{align}
where the $E(w_{j-1}^a)$ is the embedding of the previous aspect $w_{j-1}^a$. 
$\B{h}_{j}^a$ is fed into an MLP layer to obtain the probability of target aspect $w_{j}^a$ as,
\begin{equation}\label{equ:aspect_p}
\mathcal P (w_{j}^a) = \mbox{softmax}(\B{W}_a \B{h}_j^a +\B{b}_a),
\end{equation}
where $\B{W}_a\in \mathbb{R}^{d_1\times d_a}$ is a trainable parameter, $d_a$ is the size of the aspect vocabulary.
\subsubsection{Explanation Decoder}
After the user interested aspect set $w^a$ has been generated, we can further generate aspect-relevant explanations. 
For the $j$-th explanation $p_{j}\in P$, we utilize the $j$-th hidden state $\B{h}_j^a$ of aspect encoder as the initial hidden state $\B{h}_{j,0}^p$. Similar to Eq.~\eqref{equ:att_aspect}, we further use review-based syntax graph to obtain attention vector $\B{c}_{u,j,t-1}^p$ and $\B{c}_{i,j,t-1}^p$ at each decoding time-step $t-1$. We also concatenate $\B{h}_{j,0}^p$, $\B{c}_{u,j,t-1}^p$, $\B{c}_{i,j,t-1}^p$ and previous hidden state to obtain $\hat {\B{h}}_{j,t-1}^p$. Then $\hat {\B{h}}_{j,t-1}^p$ and the embedding of previous predicted word $E(w_{j,t-1}^p)$ are fed into the decoder as,
\begin{equation}\label{equ:mse}
\B{h}_{j,t}^p = \mbox{LSTM}(\hat{\B{h}}_{j,t-1}^p,E(w_{j,t-1}^p)).
\end{equation}
The probability of target word $w_{j,t}^p$ is calculated as,
\begin{equation}\label{equ:explanation_p}
\mathcal P(w_{j,t}^p) = \mbox{softmax}(\B{W}_p\B{h}_{j,t}^p +b_p),
\end{equation}
where $\B{W}_p\in \mathbb{R}^{d_1\times d_v}$ is a weight parameter and $d_v$ is the size of the  vocabulary.

\subsection{Multi-task Learning Objective Function}

For the explanation generation task, we define the following cross-entropy losses for aspect and explanation generation respectively,
\begin{align}
\ell_{ui}^{(a)} &= \frac{1}{|\C{A_{ui}}|}\sum_{j=1}^{|\C{A_{ui}}|}-\log(\mathcal P (w_{j}^a)), \nonumber\\
\ell_{ui}^{(p)} &= \frac{1}{|\C{P_{ui}}|}\sum_{j=1}^{|\C{P_{ui}}|}\frac{1}{|p_j|}\sum_{t=1}^{|p_j|}-\log(\mathcal P (w_{j,t}^p)),
\end{align}
where $|\C{A_{ui}}|$ and $|\C{P_{ui}}|$ denote the length of golden aspect and explanation sets for a given user-item pair $(u, i)$. $|p_j|$ denotes the length of the ground-truth explanation for the $j$-th aspect. $\mathcal P(w_{j}^a)$ and $\mathcal P(w_{j,t}^p)$ denote the probability of aspect $w_{j}^a$ and  word $w_{j,t}^p$. In addition, we also choose preference prediction as an auxiliary task to learn the HAG model and define the loss function as follows,
\begin{equation}\label{equ:mse}
\ell_{ui}^{(r)} =(\hat{r}_{ui} - r_{ui})^2,
\end{equation}
where $\hat{r}_{ui}$ and $r_{ui}$ denote the predicted and ground-truth rating values respectively. The final loss function of the proposed HAG model is defined as follows,
\begin{equation}\label{equ:allloss}
\frac{1}{|\C{O}|}\sum_{(u,i)\in \C{O}}\ell_{ui}^{(a)} + \ell_{ui}^{(p)} + \ell_{ui}^{(r)},
\end{equation}
where $\C{O}$ denotes the set of observed user-item pairs in the training data, and $|\C{O}|$ is the cardinality of set $\C{O}$.
The entire framework can be effectively trained by minimizing Eq.~\eqref{equ:allloss} using end-to-end back propagation.

\section{Experiments}
In this work, we perform experiments to evaluate both the explanation generation performance and preference prediction performance of the proposed HAG model.

\subsection{Experimental Settings}
\label{subsection:my}
\subsubsection{Experimental Datasets}
\label{subsubsection:dataset_details} 
The experiments are conducted on the Amazon Review dataset~\cite{he2016ups} and Yelp Challenge 2019 dataset\footnote{https://www.yelp.com/dataset/challenge}, which have been widely used for explanation generation. For the Amazon review dataset, we choose the following 5-core subsets for evaluation: "Kindle Store" and "Electronics" (respectively denoted by Kindle and Elec.). For the Yelp dataset and Amazon review dataset, we keep users and items that have more than 20 and 5 reviews for experiments respectively, due to the limitation of computation resources. In each dataset, a record consists of user ID, item ID, overall rating, and textual review. Following previous studies\cite{cheng2018aspect,he2015trirank}, we first extract aspects from the review data by the tool developed in~\cite{zhang2014explicit}. Then, we only keep records that contain more than one aspect and extract aspect-relevant sentences from reviews as target explanations. Table~\ref{tab:dataset} summarizes the statistics of experimental datasets. For each dataset, we randomly split the data into training, validation, and test data by the ratio 8:1:1.
\begin{table}
\centering
\begin{tabular}{l|cccccc}
\hline
Dataset  & \# Users & \# Items & \# Reviews & \# Aspects \\ \hline
Kindle       & 44,589  & 46,691 & 650,575  & 2,683    \\
Elec.      & 118,607  & 45,334  & 1,036,965  & 7,345     \\
Yelp      & 22,982  & 15,525  & 1,072,495  & 7,070      \\
\hline
\end{tabular}
\caption{Statistics of the experimental datasets.}
\label{tab:dataset}
\end{table}

\subsubsection{Evaluation Metrics}

For the explanation generation task, we use BLEU~\cite{papineni2002bleu}, ROUGE~\cite{lin2004rouge}, and METEOR~\cite{banerjee2005meteor} as the evaluation metrics, which evaluate the text similarity between the generated and gold explanations. BLEU evaluates the n-gram overlap between gold and generated explanations. ROUGE evaluates the recall, precision, and accuracy of the n-gram overlap. METEOR calculates the harmonic mean of each word precision and recall based on the whole corpus. However, these traditional language generation metrics can not measure whether the predicted explanations could express the gold aspects. To better evaluate the generated explanation, we also employ the Feature Matching Ratio (FMR) \cite{li2020generate} to measure whether generated explanation can include the target aspects. 

To evaluate preference prediction performance of different methods, we use Mean Absolute Error (MAE) as the evaluation metric. The definition of MAE is as follows, 
\begin{equation}\label{equ:mse}
\mbox{MAE} = \frac{1}{|\C{T}|}\sum_{(u,i)\in \C{T}}\big|\hat{r}_{ui} - r_{ui}\big|,
\end{equation}
where $\C{T}$ denotes the set of test data, $\hat{r}_{ui}$ denotes the predicted rating value, $r_{ui}$ denotes the rating value in test data, $|\cdot|$ denotes the size of a set.

Note that larger BLEU, ROUGE, METEOR, and FMR values indicate better results for explanation generation task, and lower MAE values indicate better performance for preference prediction task.

\begin{table*}
\centering
\begin{tabular}{l|l|cccccc|cccccc}
\hline
\multirow{2}{*}{Datasets} & \multirow{2}{*}{Methods}     & \multicolumn{6}{c}{Single-aspect Generation}   & \multicolumn{6}{|c}{Multi-aspect Generation} \\ 

\cline{3-14}
         &  & FMR  & B-1 (\%)   & B-4 (\%)  & R-1 (\%)   & R-L (\%)   & M (\%) & FMR  & B-1 (\%)   & B-4 (\%)  & R-1 (\%)  & R-L (\%)  & M (\%)  \\ \hline
\multirow{6}{*}{Kindle}
& Att2Seq & 0.22 & 13.96 & 1.95 & 14.96 & 11.77 & 5.81 & \underline{0.46}  & \underline{18.01}& 2.19  & 19.80 & 13.47 & \underline{6.88} \\
& ExpNet & 0.27 & 14.80  & 2.86 & 15.65 & 12.80 & 5.95 & 0.39 & 12.87  & 2.13  & 17.31  & 11.97 & 5.09 \\

& Ref2Seq & \underline{0.29}      & \underline{15.46}& {3.07}        & \underline{17.66}           & \underline{14.42}        & \underline{6.68} &  \underline{0.46}       & 13.47   & \underline{2.29}        & \underline{20.91}              & \underline{14.34}        & 6.27    \\
& NETE-PMI & 0.16 &12.09 & 1.21 & 13.20 & 10.32 & 4.96 &   -   &  -     &   -  &  - &   -   &    -   \\
& ACF & {0.13}& 14.55 & \underline{3.08}  & 15.58 & 13.14&5.96  &0.18& 12.52  & 2.20 & 17.43     &11.80 &5.20 \\
& HAG &  \textbf{0.43}   &  \textbf{16.42} & \textbf{3.34} &  \textbf{18.08} & \textbf{14.65} &  \textbf{7.15} & \textbf{0.65}& \textbf{19.10} & \textbf{2.66} & \textbf{23.13} & \textbf{14.88} & \textbf{8.37} \\
\hline
\multirow{6}{*}{Elec.}
& Att2Seq & 0.06 & 11.08 & 0.43 & 11.35 & 8.81 & 4.25 & 0.18 & \underline{14.93} & \underline{0.67} & 16.25 & 11.03& \underline{5.29} \\
& ExpNet  & 0.05 & 12.03  & 0.36 & 12.20 & 9.51 & 4.60 & {0.18} & 14.64  & 0.58 & 16.69 & 11.45 & 4.97 \\

& Ref2Seq &\underline{0.10}        & \underline{12.70}  & \textbf{0.66}        & \underline{14.05}           & \textbf{10.99}        & {4.78}   & \underline{0.21}        & 11.78  & 0.64        & \underline{17.82}            & \underline{11.98}        & 4.86   \\
& NETE-PMI & \textbf{0.11}        & 11.28   & 0.62        & 12.08        & 9.45        & \underline{4.90}     &  -     &    -   &  -     &   -    &  - &  -    \\
& ACF & 0.08 & 11.31  & 0.34 & 12.51 &9.57 &3.33 &{0.18} & 11.99 & 0.31 & 14.69 &9.38 &4.08  \\
& HAG &  \textbf{0.11}        & \textbf{13.84}   & \underline{0.65}        & \textbf{14.10}            & \underline{10.90}        & \textbf{5.44} &\textbf{ 0.31}        & \textbf{16.78} & \textbf{0.86}        & \textbf{18.54}        & \textbf{13.34}        & \textbf{6.97} \\\hline
\multirow{6}{*}{Yelp}
& Att2Seq & 0.06 & 12.06 & 0.92 & 12.45 & 9.84 & 4.33 & 0.19 & \underline{16.70} & 0.91 & 17.19 & 13.00 & \underline{5.61}\\
& ExpNet & 0.08  & 8.53 & 0.69   & 14.34   & 11.02   & 4.44  &{0.22} &16.13 & \underline{1.03}  & 18.23 & 14.47 & 5.55    \\

& Ref2Seq &{0.09}     & \underline{13.32} & \underline{1.05}      & \underline{16.16}         & \underline{12.87}     & \underline{5.15}  & \underline{0.23}        & 13.45  &0.96       & \textbf{18.93}             & \underline{14.67}        & 5.42      \\
& NETE-PMI & \underline{0.10}        & 11.73 & 0.84      & 14.22       & 11.05   & 4.40 &  -    &    -   &  -    &   -    &  - &  - \\
& ACF & 0.08 & 12.26 & 0.54 & 14.36 &11.87  &3.74 & {0.22} & 14.09 & 0.65 & 15.47 & 11.63 &4.08\\
& HAG  & \textbf{0.25}  & \textbf{17.40}  & \textbf{1.17}       & \textbf{19.05}       & \textbf{14.64}        & \textbf{7.37} & \textbf{0.27}       & \textbf{17.07}    & \textbf{1.07}      & \underline{18.30}      & \textbf{14.70}        & \textbf{6.88}  \\\hline
\end{tabular}
\caption{Explanation generation performance achieved by different methods in terms of FMR, BLEU (\%), ROUGE (\%), METEOR (\%). B, R and M refer to BLEU, ROUGE and METEOR. Note that NETE-PMI generates explanations with a single aspect, thus we only report its single-aspect explanation generation performance.}

\label{tab:textgeneration}
\end{table*}

\subsubsection{Baseline Methods}
\label{baseline_methods}
We compare HAG with the following state-of-the-art explanation generation methods, 
\begin{itemize}
    \item \textbf{Att2Seq}~\cite{dong2017learning}: It incorporates the Seq2Seq model~\cite{sutskever2014sequence} and attention mechanism to learn the user's preference from the user attributes and generate review explanations.
    \item \textbf{ExpNet}~\cite{ni2018personalized}: It utilizes an encoder-decoder framework to expand a short phrase to a long review by combining the user and item information with other auxiliary side information.
    \item \textbf{Ref2Seq}~\cite{ni2019justifying}: This method follows the structure of Seq2Seq and learns the representation from the user and item reviews to generate explanations.
    \item \textbf{NETE-PMI}~\cite{li2020generate}: It adopts MLP to predict the rating and then generates a template-controlled sentence with the predicted aspect.
    \item \textbf{ACF}~\cite{li2019generating}: It uses MLP to encode different attributes and applies a coarse-to-fine decoding model to generate long reviews.
\end{itemize}

Moreover, we also compare HAG with the following rating prediction methods to evaluate its ability in predicting users' preferences,
\begin{itemize}

    \item \textbf{PMF}~\cite{mnih2008probabilistic}: This is the probabilistic matrix factorization method developed for rating prediction.
    
    \item \textbf{SVD++}~\cite{koren2008factorization}: This method exploits both the user's preferences on items and the influences between items for recommendation.

    \item \textbf{CARL}~\cite{wu2019context}: This method uses CNNs to learn relevant aspects from the review data;
    
    \item \textbf{RMG}~\cite{wu2019reviews}: It uses a multi-view learning framework to incorporate the review contents and the users' rating behaviors for the recommendation. 
    
    \item \textbf{NETE-PMI}~\cite{li2020generate}: This method feeds user ID and Item ID to an MLP to predict rating scores.
\end{itemize}

\subsubsection{Implementation Details}
\label{implement_details}

In this work, all the evaluated methods are implemented by PyTorch~\cite{NEURIPS2019_9015}. For explanation generation methods, the learning rate is chosen from $\{0.0005, 0.0007, 0.002\}$, and the batch size is set to 16. We empirically set the max vocabulary size $d_v$ to 30,000. All the remaining words are replaced by the special token \textlangle UNK\textrangle . For single-aspect explanation generation, we set the max sentence length to 15. For multi-aspect explanation generation, we set the max feature length to 4 and set the max sentence length to 25. During the inference process, we use the greedy search algorithm to generate explanations. 

In HAG, we set the hidden size $d_1$ and word embedding size to 128. The pre-trained Google News vectors\footnote{https://code.google.com/archive/p/word2vec/} are used to initialize the word embeddings. The graph node embedding size $d_0$ is chosen from $\{8, 16, 32, 64, 128\}$, and the dimension of user and item id embeddings $d_2$ is chosen from $\{8, 16, 32, 64, 128\}$. Adam is used to optimize the model with a cosine annealing learning rate decay~\cite{huang2017snapshot}. The number of graph pooling layers $L$ is set to 2. Moreover, we set the number of top-ranked aspects $n$ extracted from the user/item reviews to 4. 
In Att2Seq,  we set the dimension of attribute embeddings to 64. For the decoding process, the dimension of word embeddings and hidden vectors are set to 512. The dropout rate is set to 0.2. For the ExpNet model, we set the dimension of attribute embeddings and word embedding to 64 and 512 respectively. The dimension of the aspect embedding is 15. For the decoder part, we set the hidden size of GRU to 512, and the dropout rate is set to 0.1. For Ref2Seq, the dimension of word embeddings and hidden vector are 256. For encoder and decoder, the dropout rates are set to 0.5 and 0.2. For NETE-PMI, we set the dimensions of word embeddings and attribute vectors to 200. The size of RNN hidden states is set to 256. The dropout rate is 0.2, and the regularization parameter is set to 0.0001.  In ACF, aspects are extracted by TwitterLDA, and the number of aspects is set to 10. For each aspect, the number of keywords is set to 50. The dimension of word embeddings is set to 512. The hidden size of the 2-layer GRU is set to 512. 

The hyper-parameters of preference prediction methods are set as follows. For PMF and SVD++, the learning rate is chosen from \{0.0001, 0.0005, 0.001, 0.005, 0.01, 0.05\}, the dimensionality of embeddings is chosen in \{8, 16, 32, 48, 64, 128\}, and the batch size is set to 512.  
For RMG, the learning rate is selected from \{0.01, 0.005, 0.001, 0.0005\}, the number of CNNs filters is set to 150 and the kernel size is 3. For CARL, the learning rate is chosen from the \{0.05, 0.01, 0.005, 0.001\}. The dimension of ID embedding and hidden state are 50 and 200. The number of CNN filters is 40.

\begin{table}
\centering
\begin{tabular}{l|ccc}
\hline
Methods   & Kindle & Electronics  & Yelp   \\ \hline
PMF      & 0.6214 & 0.8292 & 0.7982 \\
SVD++    & 0.5723 & 0.8492 & 0.8022 \\ 
CARL     & 0.5223 & \textbf{0.7794} & 0.7766 \\
RMG      & 0.5555 & 0.8203 & 0.8035 \\ 
NETE-PMI & 0.5284 & 0.8537 & \underline{0.7763} \\
HAG     & \textbf{0.5069} & \underline{0.7836} & \textbf{0.7760}\\
\hline
\end{tabular}
\caption{The preference prediction performance achieved by different methods in terms of MAE. The best results are in \textbf{bold} faces and the second-best results are \underline{underlined}.}
\label{tab:Preference prediction task}
\end{table}

\subsection{Performance Comparison}
\begin{figure*}
    \begin{subfigure}[b]{0.24\textwidth}
        \centering
        \includegraphics[width=\textwidth]{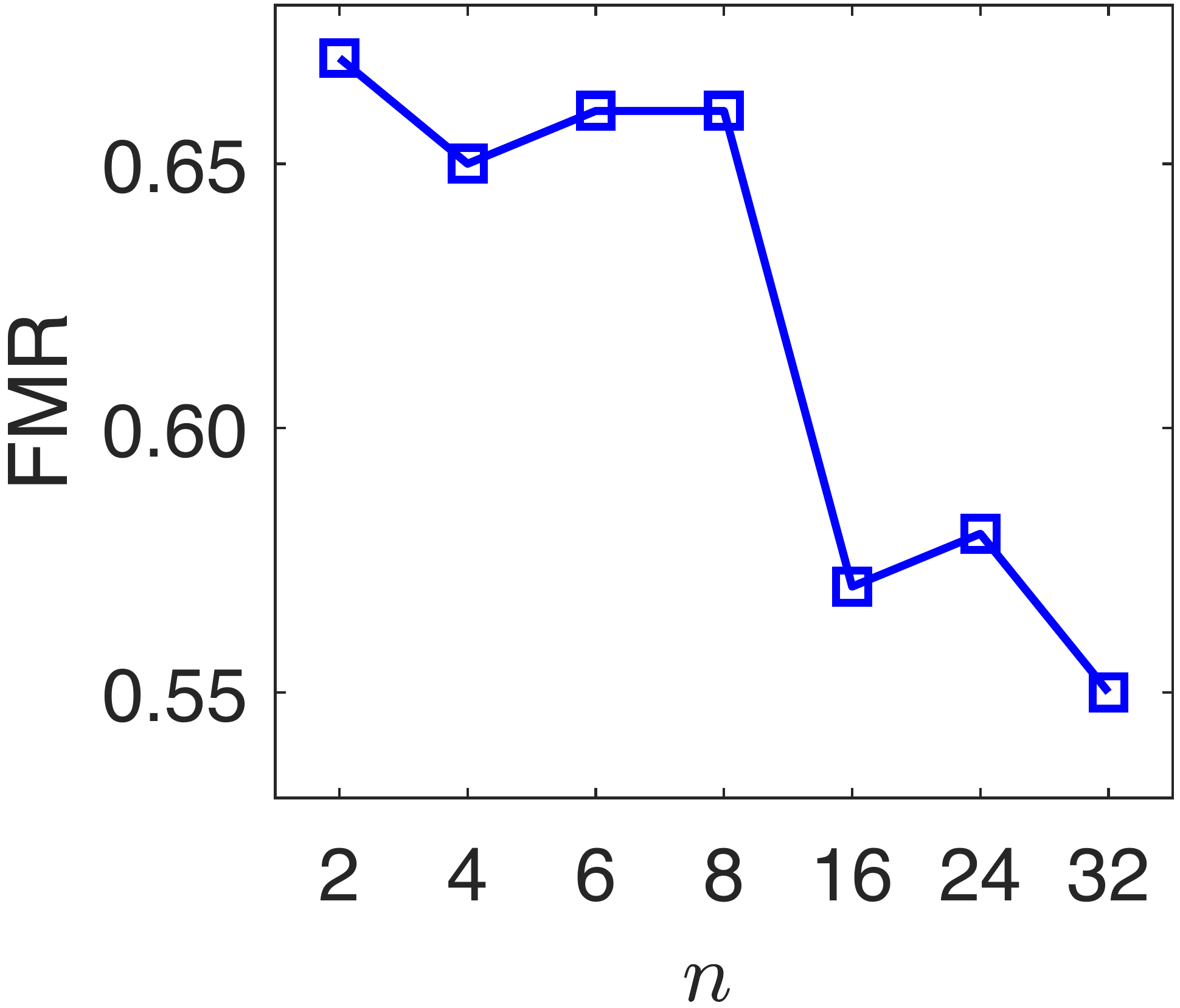}
        \caption{FMR}
    \end{subfigure}\hfill
    \begin{subfigure}[b]{0.235\textwidth}
        \centering
        \includegraphics[width=\textwidth]{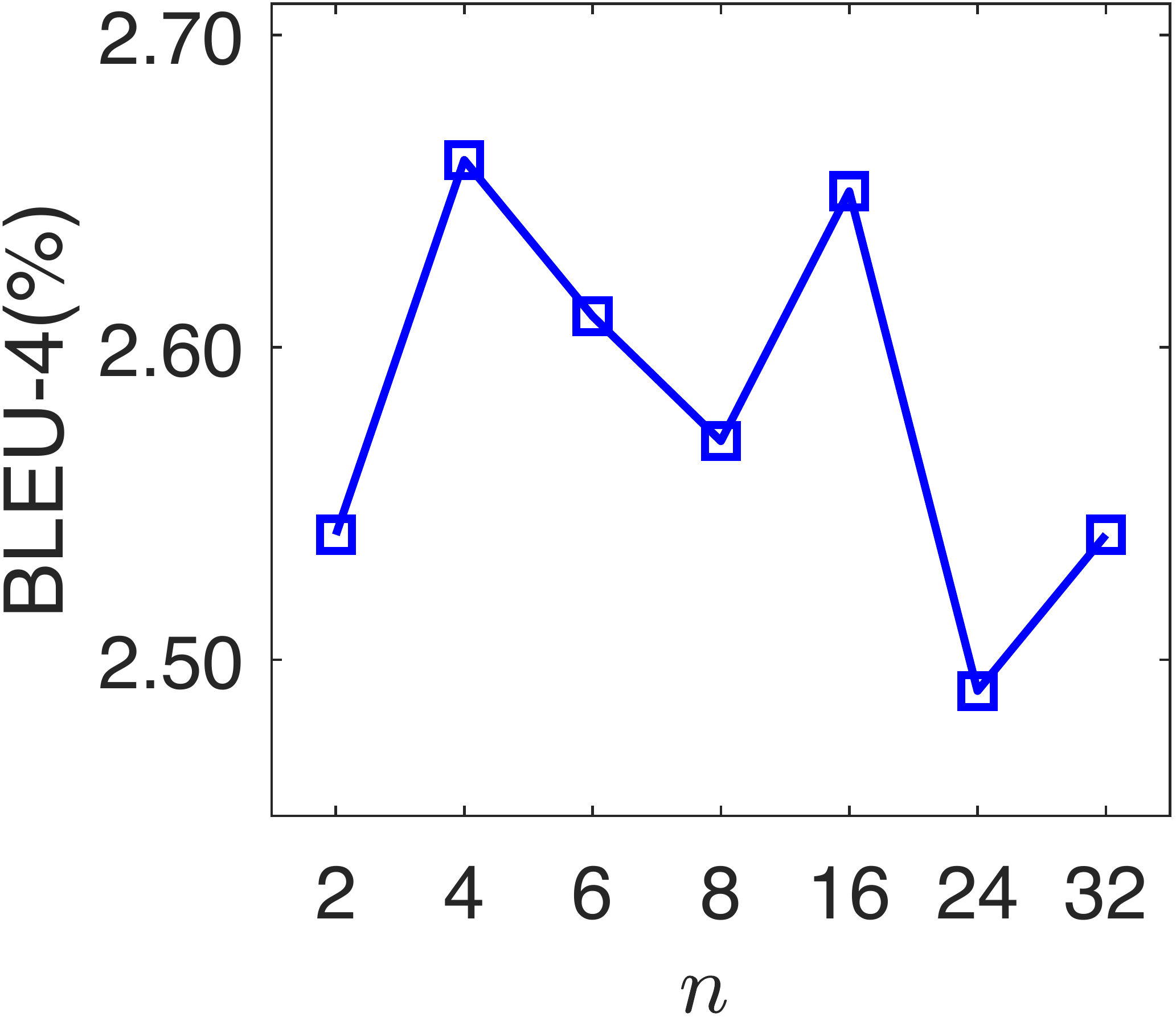}
        \caption{BLEU-4}
    \end{subfigure}\hfill
    \begin{subfigure}[b]{0.24\textwidth}
        \centering
        \includegraphics[width=\textwidth]{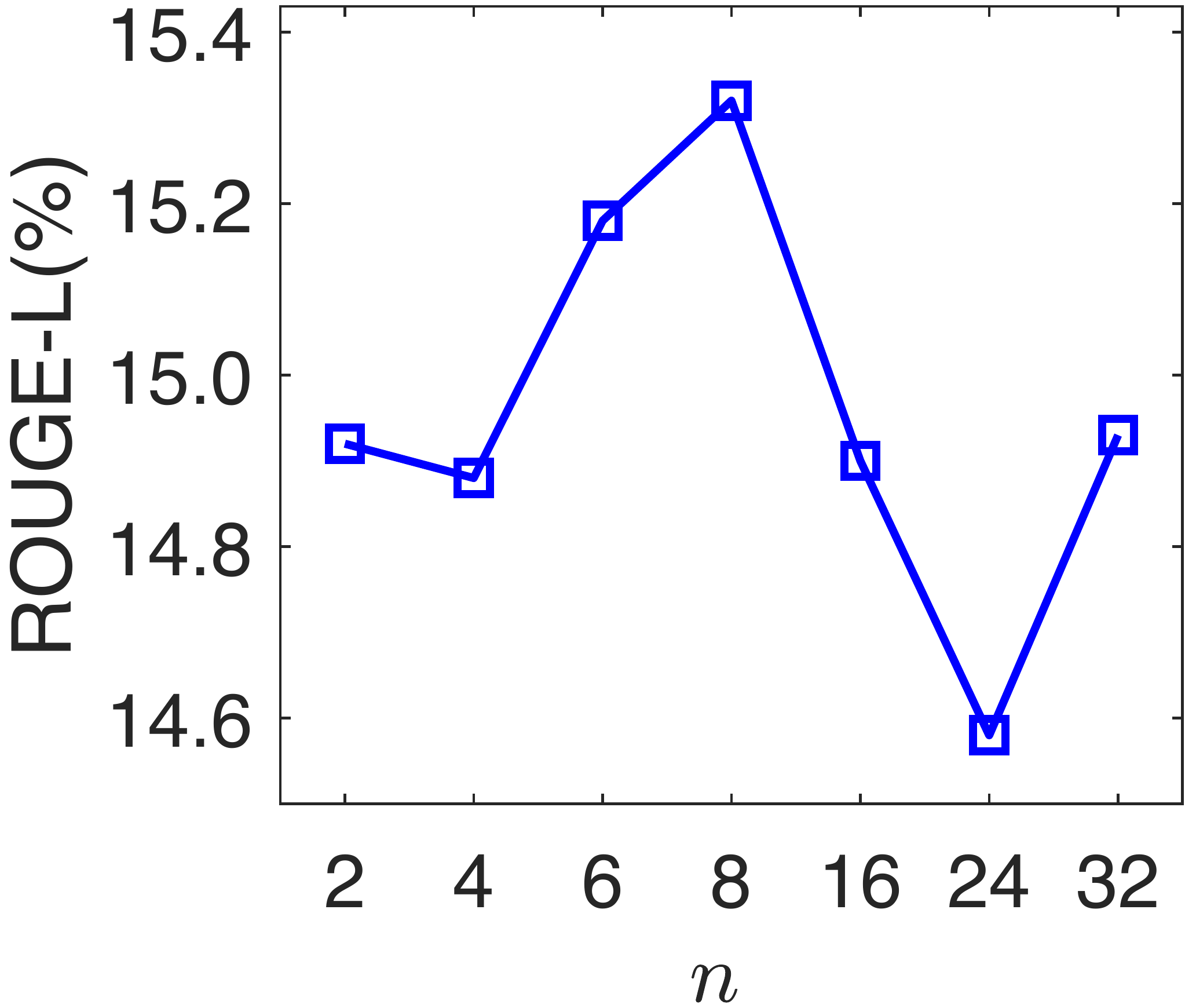}
        \caption{ROUGE-L}
    \end{subfigure}\hfill
    \begin{subfigure}[b]{0.24\textwidth}
        \centering
        \includegraphics[width=\textwidth]{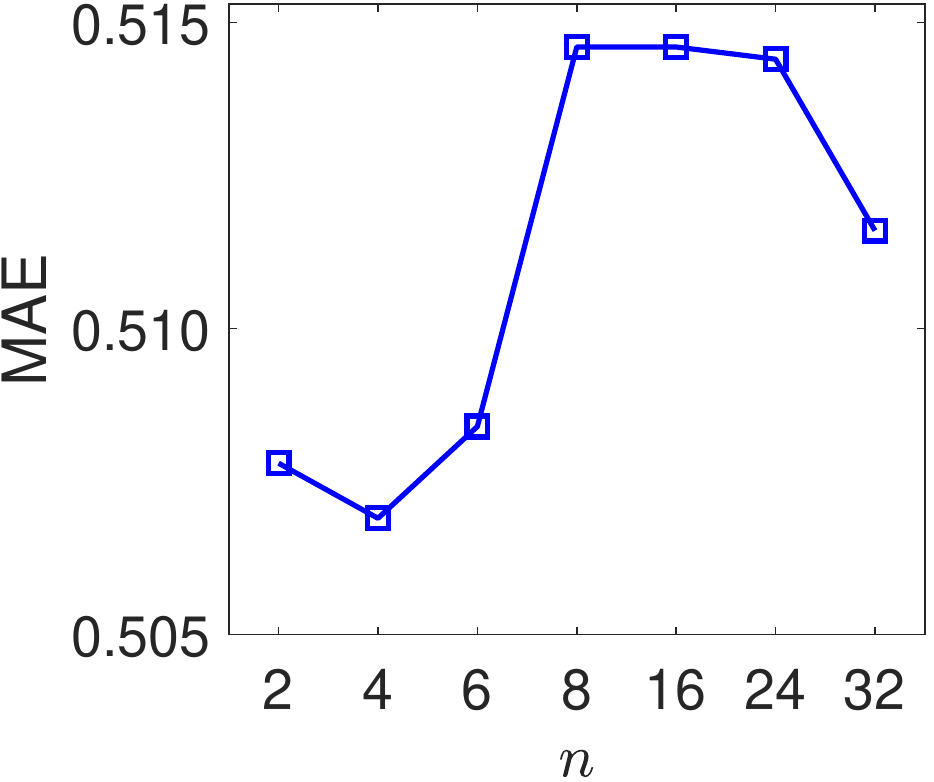}
        \caption{MAE}
    \end{subfigure}
    \caption{Performance of HAG with respect to different settings of historical aspect number $n$ on Kindle dataset.} 
    \label{fig:aspect_num}
\end{figure*}

Table~\ref{tab:textgeneration} summarizes the performance of the single-aspect and multi-aspect explanation generation tasks achieved by different methods. 
As shown in Table~\ref{tab:textgeneration}, ExpNet usually achieves better performance than Att2Seq by using aspects to guide the generation process. Ref2Seq obtains better results than ExpNet and Att2Seq. One potential reason is that Ref2Seq uses the user's historical reviews as inputs for explanation generation.
In addition, ExpNet outperforms NETE-PMI on the Kindle dataset, in terms of all metrics. It may because that ExpNet uses an attention fusion layer to control the generation outputs. Compared with Ref2Seq, ACF does not extract side information from the review data as input, and it also generates sentences by filling the generated templates based on predicted aspects. Its explanation generation performance is highly dependent on the quality of input aspects. In ACF, the aspects are extracted by TwitterLDA~\cite{zhao2011comparing}, thus the aspect quality could be influenced by the review data quality and the given number of latent topics.

The preference prediction performance achieved by different methods is shown in Table~\ref{tab:Preference prediction task}. We can note that the review-based methods usually achieve better performance than traditional matrix factorization method (\ie PMF and SVD++). 
Moreover, the proposed HAG method achieves the best preference prediction performance on Kindle and Yelp datasets, and achieves the second best prediction performance on Electronics dataset, in terms of MAE. These observations indicate that the user preference predicted by HAG can support the high-quality explanation generation. 

For both tasks, HAG usually achieves the best performance on all datasets, by enhancing explanation generation with user and item review-based syntax graphs. The review-based syntax graph aggregates information of all the reviews associated with the user/item to provide a unified view about the aspect-relevant details about the user/item. Thus, the review-based syntax graph can also help weaken the impacts caused by noise reviews.

\begin{table}
\centering
\begin{tabular}{l|ccccc}
\hline
Methods                       & FMR   & BLUE-4 (\%) & ROUGE-L (\%) & MAE       \\ \hline
HAG                        & 0.65  & \textbf{2.66}   & {14.88}   & \textbf{0.5069}    \\ \hline
HAG\textsubscript{w/o AGP} & \textbf{0.68}  & 2.39   & 14.40   & 0.5075      \\
HAG\textsubscript{GAT}     & 0.65  & 2.33   & 14.41   & 0.5099    \\
HAG\textsubscript{LSTM}    & 0.63  & 2.43   & 14.71   & 0.5127    \\\hline
HAG\textsubscript{MLP}     & 0.57  & {2.41} & 12.24    &0.5095   \\
HAG\textsubscript{implicit} & 0.64 & {2.22} & {14.90} &0.5084\\\hline
HAG\textsubscript{w/o Relation} & 0.57 & {2.58} & \textbf{15.50}  &0.5141   \\
\hline
\end{tabular}
\caption{The performance achieved by different variants of HAG on Kindle dataset.}
\label{tab:ablation}
\end{table} 

\subsection{Ablation Study}

In this section, we perform ablation study to analyze the effectiveness of different components of HAG.

\subsubsection{Impacts of AGP Operator}
To study the contributions of the AGP operator, we study the performance of the following three variants of HAG,
\begin{itemize}
    \item \textbf{HAG}\textsubscript{w/o AGP}: In this variant, the AGP operator is removed from the model. For the aspect and explanation attention, we use the node feature of the original review-based syntax graph to replace the node feature of the $L$-th graph.
    
    \item \textbf{HAG}\textsubscript{GAT}: In this variant, the aspect-aware graph pooling module of AGP is removed. We only use GAT to learn the representation of the review-based syntax graph.
    
    \item \textbf{HAG}\textsubscript{LSTM}: In this variant, the AGP operator is replaced with LSTM to update the representations of nodes in the graph. For the aspect and explanation attention calculation, it is same as \textbf{HAG}\textsubscript{w/o AGP}.
\end{itemize}

As shown in Table~\ref{tab:ablation}, HAG outperforms \textbf{HAG}\textsubscript{w/o AGP}, \textbf{HAG}\textsubscript{GAT}, and \textbf{HAG}\textsubscript{LSTM}, in terms of BLEU-4, ROUGE-L, and MAE. This demonstrates that the AGP operator can benefit both the prediction of user preferences and the explanation generation.

\subsubsection{Impacts of Historical Aspects}

To study the impacts of historical aspects, we consider the following two variants of HAG for evaluation,
\begin{itemize}
    \item \textbf{HAG}\textsubscript{MLP}: In this variant, we replace BiLSTM with MLP to extract features from the historical aspects.
    
    \item \textbf{HAG}\textsubscript{Implicit}: In this variant, we replace the feature of the historical aspect with a randomly initialized vector, which can be learned in model training. 
\end{itemize}

As shown in Table \ref{tab:ablation}, HAG achieves better performance than \textbf{HAG}\textsubscript{MLP} and \textbf{HAG}\textsubscript{Implicit}. This observation indicates that the historical aspect set is helpful to mine the user preferences for aspects, and the BiLSTM structure can help learn better representations for historical aspects. 

\subsubsection{Impacts of Grammatical Relations}

To study the impacts of grammatical relations in the review-based syntax graph, we remove the relation embedding in Eq.~\eqref{equ:dp2} and denote this variant of HAG by \textbf{HAG}\textsubscript{w/o Relation}. As shown in Table \ref{tab:ablation}, HAG achieves better FMR, BLUE-4 and MAE values than \textbf{HAG}\textsubscript{w/o Relation}. This indicates that the grammatical relations can help improve the explanation generation performance.

\begin{table}
\centering
\begin{tabular}{c|ccccc}
\hline
$\rho$   & FMR           & BLEU-4 (\%)          & ROUGE-L (\%)          & MAE \\ \hline
0.1     & \textbf{0.66} & 2.25          & 14.13        & 0.5107    \\
0.3     & {0.53}        & 2.57          & 14.67        & \textbf{0.5069}    \\
0.5     &0.65           & \textbf{2.66} & 14.88        & \textbf{0.5069}        \\
0.7     & \textbf{0.66}           & 2.49             & 14.68            & 0.5159   \\
0.9     & {0.52}           & 2.54             & \textbf{14.99}            & 0.5126   \\ \hline

\end{tabular}
\caption{Performance of HAG with respect to different graph pooling ratio $\rho$ on Kindle dataset.}
\label{tab:pool_ratio}
\end{table}

\begin{table}
\centering

\begin{tabular}{c|cccc}
\hline
      $L$      & FMR            & BLEU-4 (\%)              & ROUGE-L (\%)                            & MAE \\ \hline
1     & \textbf{0.69}  & 2.22             & 13.78             & 0.5144    \\
2     & 0.65           & \textbf{2.66}    & 14.88             & \textbf{0.5069}   \\
3     & 0.65           & 2.62             & \textbf{15.79}    & 0.5081          \\
4     & 0.68           & 2.26             & 13.70             & 0.5113    \\ \hline

\end{tabular}
\caption{Performance of HAG with respect to different settings of the number of pooling layers $L$ on Kindle dataset.}
\label{tab:graph_layer}
\end{table}

\begin{table*}[]
\centering

\resizebox{1\linewidth}{!}
{\begin{tabular}{@{}l|l|l@{}}
\hline
        & Single-aspect Explanation                                                                   & Multi-aspect  Explanation                                             \\ \hline
Reference Text &
  \begin{tabular}[c]{@{}l@{}}Good \textbf{price} and has protected my phone with no\\ complaints.\end{tabular} &
  \begin{tabular}[c]{@{}l@{}}Fairy tales and \textbf{romance} to read it. \\ The \textbf{characters} were enjoyable.\end{tabular} \\\hline \hline
Att2Seq & And charges quickly.                                                             & No complaints.                                             \\ \hline
ExpNet  & I've used this case for my ipad mini and tablet.                                 & I can't wait for the next one.                             \\ \hline
ACF &
  The could is great, the could is great. &
  I found the \textbf{characters} to be perfectly and the plot driven. \\ \hline
Ref2Seq & Nice and nice protection.                                                        & A very enjoyable read. A very good read. A very good read. \\ \hline
NETE-PMI   & I have a very good \textbf{price} and the product is very good. & -                                                          \\ \hline
HAG &
  The case is very good and the \textbf{price} is great. &
  \begin{tabular}[c]{@{}l@{}}I would recommend this book to anyone who likes a good \textbf{romance}.\\ The \textbf{characters} were interesting and the story line was good.\end{tabular} \\ \hline
\end{tabular}}
\caption{Explanations generated by different methods. The target aspects are in \textbf{bold faces}. Note that NETE-PMI cannot generate multi-aspect relevant explanations. Compared with baseline methods, the proposed HAG model can generate more informative explanations, which cover more target aspects. 
}
\label{tab:case}
\end{table*}

\subsection{Parameter Sensitivity Study}

In this work, we employ the AGP operator to mine the aspect relevant information from the review-based syntax graph. We perform experiments to evaluate the impacts of the number of historical aspects $n$ on the performance of HAG. Figure~\ref{fig:aspect_num} shows performance trends of HAG with respect to different settings of $n$. As shown in Figure~\ref{fig:aspect_num}, we can note that HAG achieves the best FMR value when $n$ is set to 2. This indicates that the generated sentences have the highest probability to cover the user's interested aspects. The best BLEU-4 and MAE values are achieved when $n$ is set to 4. Moreover, the best ROUGE-L value is achieved when $n$ is set to 8. In the experiments, we empirically set the number of historical aspects $n$ to 4, due to its best performance on BLEU-4.

Moreover, we also study the performance of HAG with respect to different settings of the graph pooling ratio $\rho$, which represents the proportion of nodes retained. When $\rho$ is set to 0.1, it means that the model only retains 10\% of nodes in each graph pooling operation. As we set the graph layer to 2, only 1\% of nodes in the review-based syntax graph are remained after two AGP operations. Table~\ref{tab:pool_ratio} summarizes the performance of HAG with respect to different settings of $\rho$. We can note that the best FMR values are achieved by setting $\rho$ to 0.1 and 0.7. For text generation metrics, the best BLEU-4 and ROUGE-L values are achieved when $\rho$ is set to 0.5 and 0.9, respectively. The best preference prediction performance in terms of MAE is achieved when $\rho$ is set to 0.3 and 0.5. 

Furthermore, we also conduct experiments to study the performance of HAG with respect to different number of graph layers $L$ on the Kindle dataset. As shown in Table~\ref{tab:graph_layer}, HAG achieves better FMR values when $L$ is set to 1 and 4. However, better BLEU-4, ROUGE-L, and MAE values are achieved when $L$ is set to 2 and 3. 
When $L=2$, HAG can achieve the best performance in terms of BLEU-4 and MAE, thus it is appropriate to set $L$ to 2 in the experiments.

\subsection{Case Study}

Table \ref{tab:case} shows examples of the single-aspect and multi-aspect explanations generated by different methods. For the single-aspect generation task, we can note that the "price" is the target aspect. As shown in Table \ref{tab:case}, the explanations generated by HAG can better express the target aspect. The explanations generated by Att2Seq and ExpNet include irrelevant properties of the product. The sentences generated by ACF and Ref2Seq are not natural enough as the outputs have repetitive words or sentences. Moreover, we can note that ACF cannot generate a meaningful explanation. The main reason is that the quality of generated explanations highly depends on the accuracy of the predicted aspects, and ACF cannot predict the user's interested aspects accurately. Although NETE-PMI can cover the target aspect, the subject of the sentence is not correct. For the multi-aspect generation task, we can note that most outputs of baseline models cannot cover the target aspects such as ``romance" and ``characters". Thus, they cannot generate high-quality explanations based on the user preferences on aspects. The explanation sentences generated by HAG not only cover specific aspects but also are more natural. Compared to baseline methods, the proposed HAG model can generate more expressive and reliable explanations.

\begin{table}
\centering
\begin{tabular}{l|cccc}
\hline
Model   & FMR   & BLEU-4 (\%) & ROUGE-L (\%)      & MAE    \\ \hline
HAG\textsubscript{P}         & {0.65}    & \textbf{2.66}    & {14.88}  &  \textbf{0.5069}   \\
HAG\textsubscript{w/o P}        & \textbf{0.67}  & {2.60}   &                                  \textbf{15.30}   &{0.5102}   \\\hline 
Ref2Seq\textsubscript{P}     &{0.46} & {2.29}  & {14.34}   &-\\
Ref2Seq\textsubscript{w/o P}     &{0.45} & {2.22}  & {14.22}  &-\\
\hline
\end{tabular}
\caption{Performance of HAG and Ref2Seq with/without pre-trained word embeddings on Kindle dataset, where ``P" means with pre-trained embedding and ``w/o P" means without pre-trained embedding. Note that Ref2Seq does not include the preference prediction module.}
\label{tab:pretrain}
\end{table}

\subsection{Impacts of Pre-trained Embeddings}

Pre-trained embeddings are obtained from models trained on large datasets, thus they usually contain rich semantic information. To evaluate the performance of HAG without pre-trained embeddings and the effectiveness of pre-trained embeddings in our tasks, we summarize the performance of HAG with/without pre-trained embeddings on the Kindle dataset in Table~\ref{tab:pretrain}. Here, we only compare HAG with Ref2Seq, which is the best baseline method in explanation generation, and it also uses reviews as inputs. As shown in Table~\ref{tab:pretrain}, the pre-trained embeddings do not help improve the performance of the proposed HAG model. The performance of HAG without pre-trained embeddings are even better than the performance of HAG with pre-trained embeddings. For the Ref2Seq model, the performance decreases without using the pre-trained embeddings. 
In conclusion, the proposed HAG model achieves excellent performance even without pre-trained embeddings, and the effectiveness of pre-trained embeddings in the studied explanation generation task is limited.

\section{Conclusion}

This paper proposes a novel explainable recommendation model, namely \underline{H}ierarchical \underline{A}spect-guided explanation \underline{G}eneration (HAG). Specifically, HAG employs a review-based syntax graph that captures the user/item details from the review data to enhance explanation generation. An aspect-guided graph pooling (AGP) operator is proposed to distill the aspect-relevant information from the review-based syntax graph. Moreover, an aspect matching mechanism is developed to match the user preferences and item properties at the aspect level. Furthermore, a hierarchical decoder is also developed to first predict the user interested aspects and then generate the aspect-relevant explanations. The experimental results on real datasets demonstrate that HGA outperforms state-of-the-art explanation generation methods.

\section{Acknowledgments}

This research is supported by the National Research Foundation, Prime Minister’s Office, Singapore under its AI Singapore Programme (AISG Award No: AISG-GC-2019-003) and under its NRF Investigatorship Programme (NRFI Award No. NRF-NRFI05-2019-0002). Any opinions, findings and conclusions or recommendations expressed in this material are those of the authors and do not reflect the views of National Research Foundation, Singapore.





\bibliographystyle{IEEEtran}
\bibliography{ref}

\end{document}